\documentclass{article}
\usepackage{float}
\restylefloat{table}
\usepackage{spconf,amsmath,graphicx}
\usepackage{tabularx}
\usepackage{amssymb}% http://ctan.org/pkg/amssymb
\usepackage{pifont}% http://ctan.org/pkg/pifont
\newcommand{\cmark}{\ding{51}}%
\newcommand{\xmark}{\ding{55}}%
\usepackage{siunitx}
\usepackage{booktabs}
\usepackage{hyperref}
\hypersetup{
colorlinks=true,
linkcolor=blue,
filecolor=magenta,
urlcolor=cyan,
pdftitle={Overleaf Example},
pdfpagemode=FullScreen,
}

\title{3D-COCO: extension of MS-COCO dataset for scene understanding and 3D reconstruction}
\name{BIDEAUX Maxence, PHE Alice, CHAOUCH Mohamed, LUVISON Bertrand, PHAM Quoc-Cuong \thanks{Funded by the European Union under Grant Agreement Number 101073951. Views and opinions expressed are however those of the author(s) only and do not necessarily reflect those of the European Union or European Research Executive Agency. Neither the European Union nor the granting authority can be held responsible for them. \newline This publication was made possible by the use of the FactoryIA supercomputer, financially supported by the Ile-De-France Regional Council.}}
\address{Université Paris-Saclay, CEA, List, F-91120, Palaiseau, France}
\begin{document}
%\ninept
%
\maketitle
\begin{abstract}

We introduce 3D-COCO, an extension of the original MS-COCO \cite{COCO} dataset providing 3D models and 2D-3D alignment annotations. 3D-COCO was designed to achieve computer vision tasks such as 3D reconstruction or image detection configurable with textual, 2D image, and 3D CAD model queries. We complete the existing MS-COCO \cite{COCO} dataset with 28K 3D models collected on ShapeNet \cite{ShapeNet} and Objaverse \cite{Objaverse}. By using an IoU-based method, we match each MS-COCO \cite{COCO} annotation with the best 3D models to provide a 2D-3D alignment. The open-source nature of 3D-COCO is a premiere that should pave the way for new research on 3D-related topics. The dataset and its source codes is available at \url{https://kalisteo.cea.fr/index.php/coco3d-object-detection-and-reconstruction/}

\end{abstract}
\section{Introduction}
\label{sec:intro}

For almost a decade, object detection has become a central topic in computer vision. This growing interest finds its roots in the new challenges of autonomous driving, crowd counting, anomaly detection, and smart video surveillance. As a result, many innovative neural networks such as Faster R-CNN \cite{RCNN}, YOLO \cite{YOLO}, SSD \cite{SSD}, and DETR \cite{DETR} have been developed over the years. Most of these architectures’ performances are evaluated and compared through some widespread datasets like Pascal VOC \cite{PascalVoc}, Open Images \cite{OpenImages} and MS-COCO \cite{COCO}.

Iterations were carried out to improve these architectures, thus making it possible to achieve optimal performance for the detection of objects appearing during training. A new area of research involves the detection of new semantic classes that do not appear during training. This innovation would allow object detectors to meet a wider field of application without the need for re-training. As an example, OV-DETR \cite{OVDETR} uses a foundation model as a backbone to transform the DETR \cite{DETR} architecture into an open vocabulary detector configurable with text and images. Then, it could be interesting to develop detection networks configurable with 3D models, but traditional detection datasets do not include 3D modality.

At the same time, 3D reconstruction approaches based on neural networks have been developed. These architectures can be used, for example, in industrial or virtual reality applications. Recently, promising 3D reconstruction methods, such as 3D-C2FT \cite{3DC2FT}, LegoFormer \cite{LegoFormer} or VPFusion \cite{VPFusion} hav emerged. Their performances are usually evaluated on ShapeNet \cite{ShapeNet}. Although this dataset includes a wide range of commonly encountered objects, it could be supplemented with new semantic classes present in detection datasets such as MS-COCO \cite{COCO}. In addition, ShapeNet \cite{ShapeNet} only offers synthetic renderings, which limits the application of 3D reconstruction networks to real-world situations.

Thus, we propose 3D-COCO, an extension of the widely used MS-COCO \cite{COCO} dataset, adapted for object detection configurable with text, 2D images, or 3D CAD (Computer-Aided Design) model queries and for single or multi-view 3D reconstruction. The 3D-COCO dataset opens new perspectives to image detection by providing 3D models that are automatically aligned with 2D annotations. It also opens the way to the integration of real images for 3D reconstruction which until now remained confined to synthetic images. Moreover, 3D-COCO provides a wider variety of semantic classes for 3D reconstruction in addition to ShapeNet's ones \cite{ShapeNet}. The Objaverse \cite{Objaverse} 3D models database is used with ShapeNet \cite{ShapeNet} to provide a sufficient number of objects for each of the 80 MS-COCO \cite{COCO} semantic classes. Alignment between collected 3D models and MS-COCO \cite{COCO} 2D annotations is made using a simple, yet effective automatic class-driven retrieval method based on IoU.

\begin{table*}[ht]
\centering
\caption{Properties of different detection and 3D reconstruction datasets. The different columns correspond to the dataset name, the number of semantic classes, the number of semantic classes shared with MS COCO \cite{COCO}, the number of images, the number of detection annotations, the number of 3D models, the synthetic or realistic nature of images, the availability of 3D models, the possibility to use the dataset for detection configurable with textual or image queries and the possibility to use the dataset for 3D reconstruction}
\label{tab:previous}
\begin{tabularx}{\textwidth} { 
  | >{\centering\arraybackslash}p{2.8cm}
  | >{\centering\arraybackslash}p{1.4cm}
  | >{\centering\arraybackslash}p{1.4cm}
  | >{\centering\arraybackslash}p{1.4cm}
  | >{\centering\arraybackslash}p{1.4cm}
  | >{\centering\arraybackslash}p{1.4cm}
  | >{\centering\arraybackslash}X
  | >{\centering\arraybackslash}X 
  | >{\centering\arraybackslash}X 
  | >{\centering\arraybackslash}X | }
 \hline
 \textbf{Dataset} & \textbf{\# Classes} & \textbf{\# COCO} & \textbf{\# Images} & \textbf{\# Ann} & \textbf{\# Models} & \textbf{Real} & \textbf{3D} & \textbf{Det} & \textbf{Rec}\\
 \hline
 COCO 2017 \cite{COCO} & 80  & 80 & 164K & 897K & N/A & \cmark & \xmark & \cmark & \xmark \\
 3DObject \cite{3DObject}  & 10  & 7 & 6.7K & N/C & N/A & \cmark & \xmark & \cmark & \xmark \\
 EPFL Car \cite{EPFLCars} & 1  & 1 & 2.3K & N/C & N/A & \cmark & \xmark & \cmark & \xmark \\
 NYU Depth \cite{NYUDepth} & 894  & N/C & 1.4K & 35.1K & N/A & \cmark & \xmark & \cmark & \xmark \\
 SUN RGB-D \cite{SUNRGBD} & $\approx$800 & N/C & 10.3K & 211.2K & N/A & \cmark & \xmark & \cmark & \xmark \\
 KITTI \cite{KITTI} & 2  & 2 & 15K & 200K & N/A & \cmark & \xmark & \cmark & \xmark \\
 IKEA \cite{IKEA} & 11  & 5 & 0.8K & 72K & 0.2K & \cmark & \cmark & \cmark & \cmark \\
 PASCAL3D+ \cite{PASCAL3D} & 12  & 12 & 30.9K & 13.8K & 0.1K & \cmark & \cmark & \cmark & \cmark \\
 ObjectNet3D \cite{ObjectNet3D} & 100  & $\approx$40 & 90.1K & 204.6K & 44.1K & \cmark & \cmark & \cmark & \cmark\\
 ABO \cite{ABO} & 63  & $\approx$15 & 398.2K & 6.3K & 8K & \cmark & \cmark & \cmark & \cmark\\
 ShapeNetCore \cite{ShapeNet} & 55  & 27 & N/A & N/A & 51.3K & \xmark & \cmark & \xmark & \cmark \\
 3D-Future \cite{3DFuture} & 34 & $\approx$5 & 20.2K & 37.4K & 10.0K  & \xmark & \cmark & \cmark & \cmark\\
 Google Scans \cite{GoogleScans} & $\approx$17  & $\approx$4 & N/A & N/A & 1K & \xmark & \cmark & \cmark & \cmark\\
 CO3D \cite{CO3D} & 50  & 50 & 1,500K & 18.6K & 18.6K & \cmark & \xmark & \cmark & \cmark\\
 Pix3D \cite{Pix3D} & 9  & 4 & 10.1K & 10.1K & 0.4K & \cmark & \cmark & \cmark & \cmark\\
 PhotoShape \cite{PhotoShape} & 1  & 1 & N/A & N/A & 5.8K & \xmark & \cmark & \xmark & \cmark \\
 Objaverse \cite{Objaverse} & N/A  & 80 & N/A & N/A & $\approx$800K & \xmark & \cmark & \xmark & \cmark \\
 Objaverse XL \cite{ObjaverseXL} & N/A & 80 & N/A & N/A & $\approx$10,000K & \xmark & \cmark & \xmark & \cmark \\
 \textbf{3D-COCO (Ours)} & 80  & 80 & 164K & 897K & 28K  & \cmark & \cmark & \cmark & \cmark \\
\hline
\end{tabularx}
\end{table*}

To summarize, we make the following contributions:
\begin{itemize}
    \item We propose 3D-COCO, a dataset adapted both to 2D-to-3D configurable detection and single or multi-view 3D reconstruction. This dataset uses as a basis the original MS-COCO \cite{COCO} detection dataset and extends it with 3D models collected from ShapeNet \cite{ShapeNet} and Objaverse \cite{Objaverse}.

\item We present an automatic class-driven method based on IoU retrieval to match each MS-COCO \cite{COCO} 2D annotation with the best 3D model in the dataset in terms of shape and geometry similarity.

\end{itemize}

\section{Related work}
\label{sec:previously}

In computer vision, combining image modalities with 3D presents significant interest due to its potential to enhance the accuracy of scene understanding and generation tasks. By integrating these complementary modalities, computer vision systems gain improved spatial awareness and object recognition capabilities, effectively addressing challenges such as occlusions, variable lighting, and perspective distortions, which are commonly encountered in analyses based on 2D images.

In the context of object detection configurable with queries, many studies have already been led to propose datasets. We can for example cite MS-COCO \cite{COCO}, 3DObject \cite{3DObject}, EPFL Car \cite{EPFLCars}, or NYU Depth \cite{NYUDepth}. Indeed, these datasets provide images with annotation files containing bounding boxes and labels which can be used for simple detection tasks or detection tasks with text queries and 2D image queries respectively extracted from labels and bounding boxes. Some other detection datasets also provide 3D CAD models such as ObjectNet3D \cite{ObjectNet3D}, ABO \cite{ABO}, etc.

Meanwhile, other datasets for 3D reconstruction tasks have been proposed, such as ShapeNet \cite{ShapeNet},  PASCAL3D+ \cite{PASCAL3D}, or more recently the extensive databases Objaverse \cite{Objaverse}, and ObjaverseXL \cite{ObjaverseXL}. 

Among all these datasets, 3D models can be provided in many different formats : multi-view images in KITTI \cite{KITTI} ; RGB-D images in SUN-RGBD \cite{SUNRGBD} ; point clouds in Google Scans \cite{GoogleScans} and CO3D \cite{CO3D} ; meshes in IKEA \cite{IKEA}, PASCAL3D+ \cite{PASCAL3D}, ObjectNet3D \cite{ObjectNet3D}, ABO \cite{ABO}, 3DFuture \cite{3DFuture}, Pix3D \cite{Pix3D} and PhotoShape \cite{PhotoShape} ; voxel grids in ShapeNet \cite{ShapeNet}.

% \begin{itemize}
%     \item Multi-view images in KITTI \cite{KITTI}
%     \item RGB-D images in SUN-RGBD \cite{SUNRGBD}
%     \item Point clouds in Google Scans \cite{GoogleScans} and CO3D \cite{CO3D}
%     \item Meshes in IKEA \cite{IKEA}, PASCAL3D+ \cite{PASCAL3D}, ObjectNet3D \cite{ObjectNet3D}, ABO \cite{ABO}, 3DFuture \cite{3DFuture}, Pix3D \cite{Pix3D} and PhotoShape \cite{PhotoShape}
%     \item Voxel grids in ShapeNet \cite{ShapeNet}
% \end{itemize}

\begin{table*}[h!]
\begin{tabular}{lccccc|lccccc} \toprule
    \textbf{COCO Label} & \textbf{ID} & \textbf{Syn ID} & \textbf{$\#$ Models} & \textbf{SN} & \textbf{OV} & \textbf{COCO Label} & \textbf{ID} & \textbf{Syn ID} & \textbf{$\#$ Models} & \textbf{SN} & \textbf{OV} \\ \midrule
    person & 1 & 05224944 & 12 &  & \cmark & wine glass & 46 & 03443167 & 59 &  & \cmark\\
    bicycle & 2 & 02834778 & 37 &  & \cmark & cup & 47 & 03152175 & 25 &  & \cmark\\
    car & 3 & 02958343 & 3514 & \cmark &  & fork & 48 & 03388794 & 20 &  & \cmark\\
    motorcycle & 4 & 03790512 & 337 & \cmark &  & knife & 49 & 03624134 & 424 & \cmark & \\
    airplane & 5 & 02691156 & 4045 & \cmark &  & spoon & 50 & 04291140 & 8 &  & \cmark\\
    bus & 6 & 02924116 & 939 & \cmark &  & bowl & 51 & 02880940 & 185 & \cmark & \\
    train & 7 & 04468005 & 389 & \cmark &  & banana & 52 & 07769568 & 52 &  & \cmark\\
    truck & 8 & 04497386 & 20 &  & \cmark & apple & 53 & 07755101 & 51 &  & \cmark\\
    boat & 9 & 02861626 & 22 &  & \cmark & sandwich & 54 & 07711710 & 16 &  & \cmark\\
    traffic light & 10 & 06887235 & 36 &  & \cmark & orange & 55 & 07763583 & 13 &  & \cmark\\
    fire hydrant & 11 & 03351744 & 82 &  & \cmark & broccoli & 56 & 07730735 & 13 &  & \cmark\\
    stop sign & 13 & 04224949 & 31 &  & \cmark & carrot & 57 & 07746183 & 12 &  & \cmark\\
    parking meter & 14 & 03897029 & 11 &  & \cmark & hot dog & 58 & 07713282 & 19 &  & \cmark\\
    bench & 15 & 02828884 & 1776 & \cmark &  & pizza & 59 & 07889783 & 44 &  & \cmark\\
    bird & 16 & 01505702 & 24 &  & \cmark & donut & 60 & 07654678 & 37 &  & \cmark\\
    cat & 17 & 02124272 & 27 &  & \cmark & cake & 61 & 07644479 & 27 &  & \cmark\\
    dog & 18 & 02086723 & 28 &  & \cmark & chair & 62 & 03001627 & 6778 & \cmark & \\
    horse & 19 & 02377103 & 22 &  & \cmark & couch & 63 & 04256520 & 3027 & \cmark & \\
    sheep & 20 & 02414351 & 6 &  & \cmark & potted plant & 64 & 00017402 & 31 &  & \cmark\\
    cow & 21 & 01890428 & 20 &  & \cmark & bed & 65 & 02818832 & 219 & \cmark & \\
    elephant & 22 & 02506148 & 34 &  & \cmark & dining table & 67 & 03205892 & 32 &  & \cmark\\
    bear & 23 & 02134305 & 14 &  & \cmark & toilet & 70 & 04453655 & 81 &  & \cmark\\
    zebra & 24 & 02393701 & 7 &  & \cmark & tv & 72 & 03211117 & 1093 & \cmark & \\
    giraffe & 25 & 02441664 & 33 &  & \cmark & laptop & 73 & 03642806 & 451 & \cmark & \\
    backpack & 27 & 02772753 & 24 &  & \cmark & mouse & 74 & 03799022 & 25 &  & \cmark\\
    umbrella & 28 & 04514450 & 21 &  & \cmark & remote & 75 & 04074963 & 66 & \cmark & \\
    handbag & 31 & 02777157 & 25 &  & \cmark & keyboard & 76 & 03085013 & 64 & \cmark & \\
    tie & 32 & 03821155 & 5 &  & \cmark & cell phone & 77 & 02992529 & 830 & \cmark & \\
    suitcase & 33 & 02773838 & 83 & \cmark &  & microwave & 78 & 03761084 & 152 & \cmark & \\
    frisbee & 34 & 03402783 & 7 &  & \cmark & oven & 79 & 03868196 & 21 &  & \cmark\\
    skis & 35 & 04235116 & 8 &  & \cmark & toaster & 80 & 04449446 & 9 &  & \cmark\\
    snowboard & 36 & 04258901 & 7 &  & \cmark & sink & 81 & 04230655 & 19 &  & \cmark\\
    sports ball & 37 & 02781674 & 115 &  & \cmark & refrigerator & 82 & 04077839 & 40 &  & \cmark\\
    kite & 38 & 03626682 & 7 &  & \cmark & book & 84 & 02873453 & 17 &  & \cmark\\
    baseball bat & 39 & 02802334 & 34 &  & \cmark & clock & 85 & 03046257 & 651 & \cmark & \\
    baseball glove & 40 & 02803372 & 10 &  & \cmark & vase & 86 & 03593526 & 596 & \cmark & \\
    skateboard & 41 & 04225987 & 152 & \cmark &  & scissors & 87 & 04155119 & 24 &  & \cmark\\
    surfboard & 42 & 04370646 & 15 &  & \cmark & teddy bear & 88 & 04406517 & 24 &  & \cmark\\
    tennis racket & 43 & 04416941 & 18 &  & \cmark & hair drier & 89 & 03488399 & 12 &  & \cmark\\
    bottle & 44 & 02876657 & 483 & \cmark &  & toothbrush & 90 & 04460427 & 13 &  & \cmark\\
    \bottomrule
\end{tabular}
\label{tab:coco_classes}
\caption{Information about MS-COCO \cite{COCO} labels (COCO Label), their ID in MS-COCO \cite{COCO}, their WordNet ID (Syn ID), the number of collected models ($\#$ Models) and their source (ShapeNet \cite{ShapeNet} (SN) or Objaverse \cite{Objaverse} (OV))}
\end{table*}

Moreover, it can be noticed that datasets represent either widespread concepts such as MS COCO \cite{COCO} or ObjectNet3D \cite{ObjectNet3D} or very specialized categories of objects such as EPFL Car \cite{EPFLCars} or KITTI \cite{KITTI}.

Some relevant information about all these datasets is summarized in Table \ref{tab:previous}. The motivation behind 3D-COCO is to provide a common object dataset that addresses most of the scene understanding and 3D reconstruction tasks. To reach such a goal, the MS-COCO \cite{COCO} detection dataset is used as a baseline. Indeed, this dataset provides 164K realistic images with many detection annotations (about 897K) representing instances of 80 semantic classes. Moreover, this dataset is used as a reference for detection, segmentation, and pose estimation tasks.

The 3D-COCO dataset is equivalent in format to the ObjetNet3D \cite{ObjectNet3D} as it provides all the necessary data to train classical detection networks, with 3D models and a 2D-3D alignment to integrate 3D in detection.  ShapeNet Core \cite{ShapeNet}, has been used to collect CAD models for the 22 classes shared with MS COCO \cite{COCO}. The collection of the remaining 58 classes has been made possible thanks to the very large diversity of semantic classes in Objaverse \cite{Objaverse}.

Finally, to address a wide variety of applications, 3D-COCO provides 3D CAD models in multiple formats. Thus, these normalized inputs will be common to every user.

\begin{figure*}[h]
    \centering
    \includegraphics[width=0.85\textwidth]{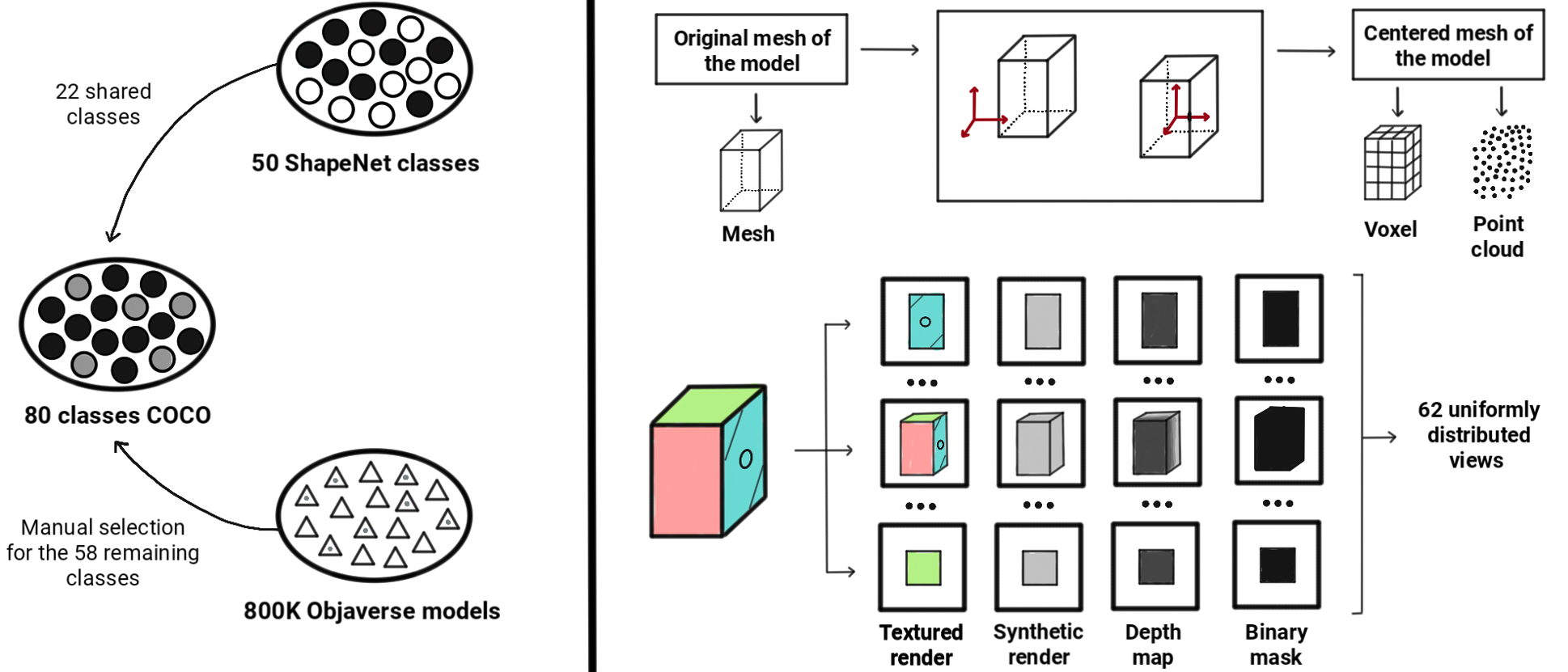}
    \caption{3D CAD models data collection from ShapeNet \cite{ShapeNet} and Objaverse \cite{Objaverse} (left) and pre-processing steps including centering, conversion (upper right), and 2D rendering (lower right). Textured renderings display the model with colors and textures, synthetic renderings display the model in a uniform gray color, depth map renderings display the nearest model points in darker colors and binary mask renderings display the silhouette of the model.}
    \label{fig:collect_and_process}
\end{figure*}

\section{Dataset creation method}
\label{sec:Methods}

\subsection{Collection of 3D models}
\label{sec:Collection}

First, 3D models are collected to create the extension of MS-COCO \cite{COCO}, which does not contain any CAD model. As represented on the left side of Figure \ref{fig:collect_and_process}, the 80 MS-COCO \cite{COCO} labels are first matched with the 55 ShapeNet \cite{ShapeNet} labels. Indeed, ShapeNet \cite{ShapeNet} provides a wide variety of common objects' good-quality models, which is important for our new dataset. The 22 matching categories, represented by the dark circles of Figure \ref{fig:collect_and_process} constitute the first contribution to 3D-COCO with 26,254 models provided.

The 58 remaining labels are then completed using the Objaverse \cite{Objaverse} 3D database. Indeed, as previously mentioned, Objaverse \cite{Objaverse} provides about 800K CAD models from a very important number of semantic classes. Thus, a manual selection is processed on the Objaverse website to complete 3D-COCO with relevant models. First, the label name is used to make the research on Objaverse \cite{Objaverse} website. Then, the universal identifier (UID) of manually selected models is stored and used later with the Objaverse \cite{Objaverse} python API to collect selected models in GLB format. Finally, collected 3D models are stored using a folder architecture described later in this paper.

As a result of the manual model collection, Objaverse \cite{Objaverse} provided 1,506 models to 3D-COCO. Figure \ref{fig:collect_and_process} illustrated the following methodology. All the information about the MS-COCO \cite{COCO} semantic classes, their identifiers, and their models are summarized in Table \ref{tab:coco_classes}.

Once collected, 3D meshes collected on Objaverse \cite{Objaverse} are converted from GLB into OBJ format using the trimesh python module to match the ShapeNet meshes format. Then, for each model of 3D-COCO, a centering operation is made by calculating its vertices' mean where each vertex coordinate is weighted by the sum of the faces containing this vertex. Following this operation,  models are pre-processed to make them available in the following formats: voxels of size 32, point clouds with 10,000 elements and 62-view $224\times224$ rendering images of different natures (textured, grey synthetic, depth map, and binary). Point clouds and voxels are generated using respectively open3d\footnote{https://pypi.org/project/open3d/} and binvox\footnote{https://www.patrickmin.com/binvox/} python modules. Rendering views are generated for each of the 4 render types by using Blender's Python API \footnote{https://pypi.org/project/bpy/}. The 62 rendering views are uniformly sampled in an Isdyakis triacontahedron structure (composed of 62 vertices). The images and voxel sizes were chosen to fit the sizes handled by most of the 3D reconstruction networks. These operations, which make the dataset usable in a wider range of applications, are illustrated on the right part of Figure \ref{fig:collect_and_process}.

\begin{figure*}[h]
\centering
\includegraphics[width=0.95\textwidth]{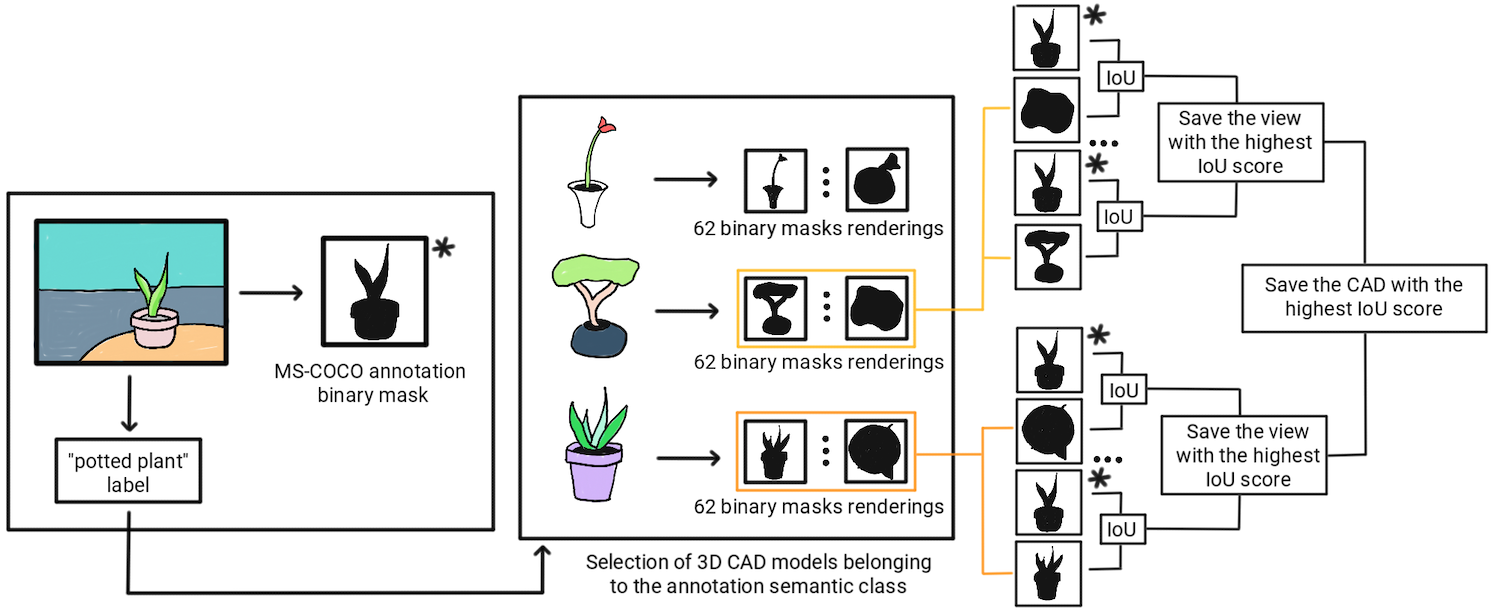}
\caption{Matching between MS-COCO \cite{COCO} 2D annotations and 3D models using our automatic class-driven retrieval method. First, the MS-COCO \cite{COCO} annotation binary mask image is extracted from the image (left). Then, the annotation label is used to select the 3D models of the same class and their 62 binary masks (middle). Finally, IoU is calculated between the MS-COCO \cite{COCO} annotation mask and each CAD binary mask: models with the highest IoU score are saved as the best matching models in the 3D-COCO annotation file} 
\label{fig:iou_matching}
\end{figure*}

\subsection{2D-3D matching}

Then, a 2D-3D alignment is implemented based on an automatic class-driven retrieval method using IoU. Indeed, CAD models contain an important amount of information about shape, and IoU is very efficient in quantifying shape similarity between elements: thus, this metric is relevant to address the 2D-3D alignment task 3D-COCO needs. In that way, each MS-COCO \cite{COCO} annotation is matched with the most representative 3D CAD models of 3D-COCO in terms of geometry and shape.

The IoU-based matching method described in Figure \ref{fig:iou_matching} requires some pre-processing both on MS-COCO \cite{COCO} annotations and on 3D-COCO models. Indeed, MS-COCO \cite{COCO} annotations and API are used to generate a binary mask for each annotation as illustrated on the left part of Figure \ref{fig:iou_matching}. This mask is saved on a $224\times224$ image and normalized in scale to make the represented element touch the edges of the picture. For each CAD model, we used the process described in \ref{sec:Collection} to get 62 binary masks represented on the bottom right side of Figure \ref{fig:collect_and_process}. This process allows to get silhouettes on images of similar sizes which will be compatible with an IoU calculation.

As illustrated on the right side of Figure \ref{fig:iou_matching}, for each MS-COCO \cite{COCO} annotation, IoU is computed between the binary mask of the annotation and the binary mask of all the render views of the 3D models sharing the same label. The best matching model is the one that provides the highest IoU. Thus, each MS-COCO \cite{COCO} annotation is matched with its 3 most representative models of 3D-COCO.

\subsection{Specific issues addressed in the annotation process}
\label{sec:difficulty}

\begin{figure}[h]
\centering
\includegraphics[width=1.0\columnwidth]{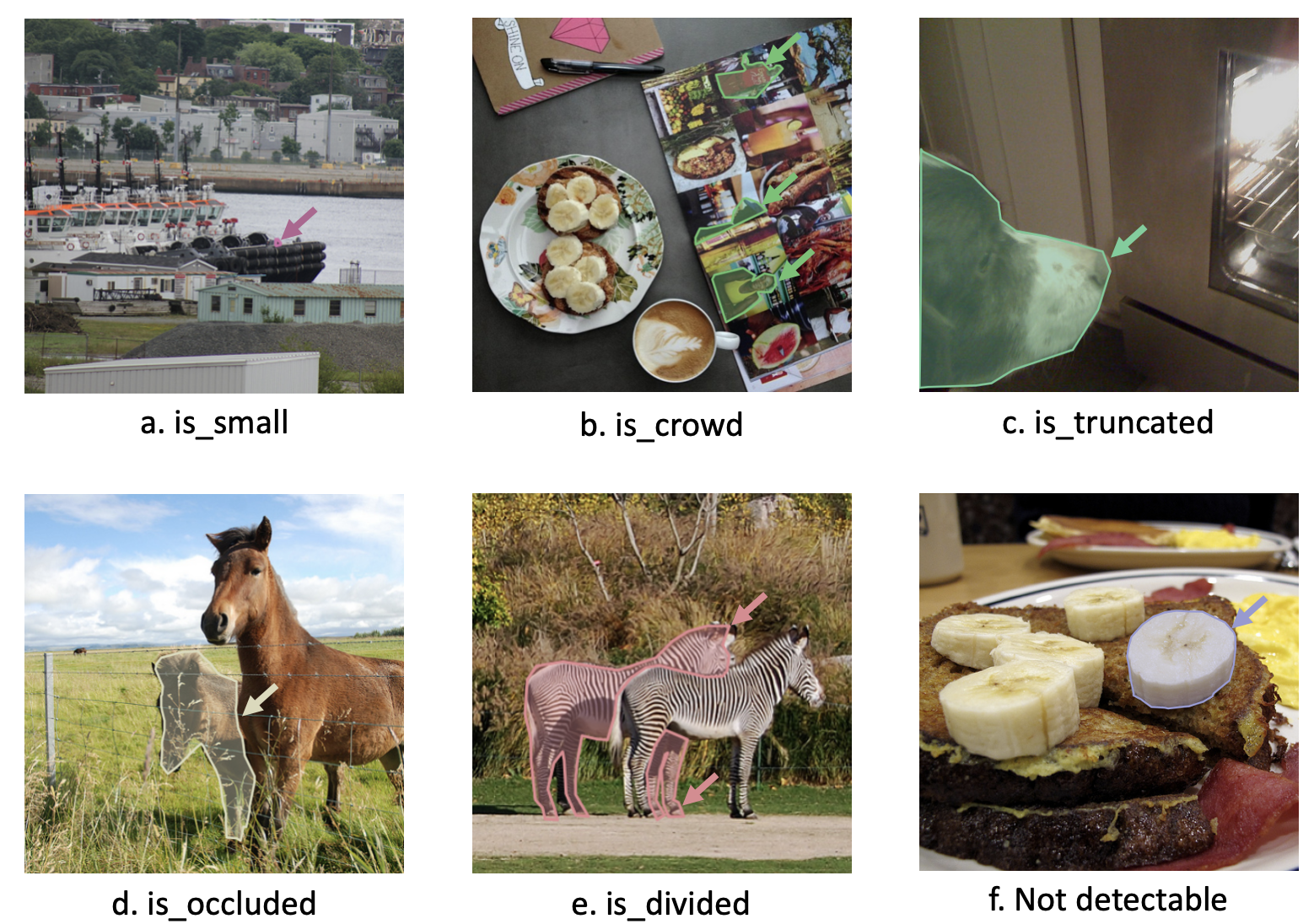}
\caption{Example of difficult scenarios from COCO \cite{COCO} images and annotations with their associated flag names}
\label{fig:anomalies}
\end{figure}

When observing MS-COCO \cite{COCO} images and annotations, some situations may harm retrieval operations or during object detection:

\begin{itemize}
    \item The annotation is too small (Figure \ref{fig:anomalies}a). This scenario is detected if the ratio between the number of pixels in the bounding box and the number of pixels in the image stays below a threshold (here $1\%$). Then, the annotation is flagged as $is\_small$.
    \item The annotation is composed of several instances appearing in a single annotation (Figure \ref{fig:anomalies}b). This scenario is detected by using the existing MS-COCO \cite{COCO} flag $is\_crowd$.
    \item The annotation is truncated (Figure \ref{fig:anomalies}c). This scenario is detected if the ratio between the distance separating the bounding box from the image edges and the image dimensions stays below a threshold (here $2\%$). Then, the annotation is flagged as $is\_truncated$.
    \item The annotation is occluded by another annotation of the image (Figure \ref{fig:anomalies}d). This scenario is detected if the annotation mask intersects another mask in the image, which results in an IoU score different from 0. Then, the annotation is flagged as $is\_occluded$.
    \item The instance is divided into multiple separated areas, (Figure \ref{fig:anomalies}e). This scenario is detected by connecting components labeling over the binary mask of the instance. Considering that each pixel shares the same label as the pixels it is connected to if more than one label appears after applying the method, the annotation is flagged as $is\_divided$.
    \item There is a lack of accuracy in the labeling of MS-COCO \cite{COCO} images and a lack of diversity in the collected 3D models (Figure \ref{fig:anomalies}f). In this example, the instance is labeled as "banana", but all the 3D models with this label represent an entire banana. This instance should then be labeled "piece of banana" or the CAD models database should be completed with meshes representing pieces of banana. This scenario is difficult to determine automatically.
    \end{itemize}

\begin{figure}[ht]
\centering
\includegraphics[width=0.9\columnwidth]{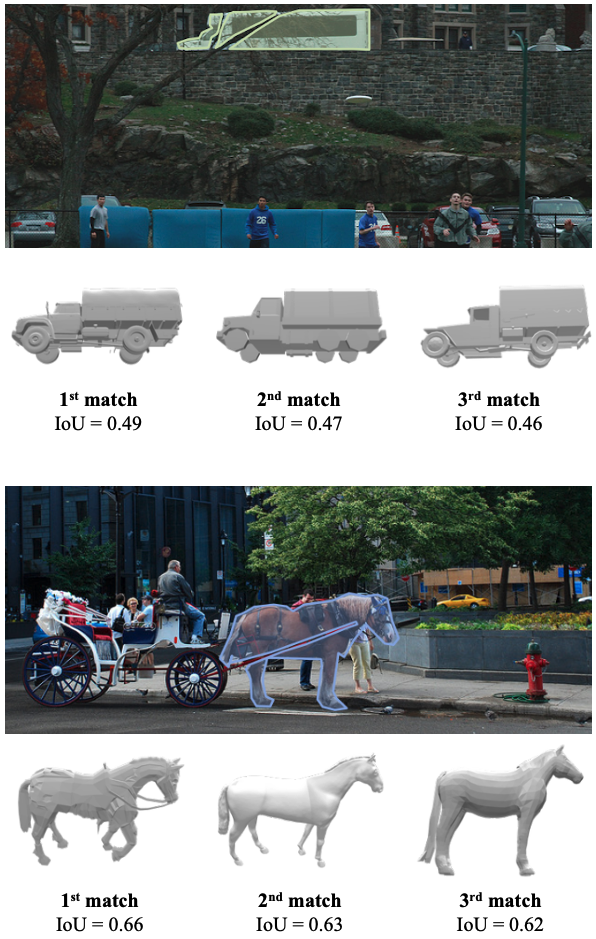}
\caption{MS-COCO \cite{COCO} images with \textit{Truck} and \textit{Horse} annotations followed by their 3 best matching models predicted by our IoU-based retrieval method}
\label{fig:retrieval}
\end{figure}

Two examples of automatic class-driven IoU-based retrieval are presented in Figure \ref{fig:retrieval}.

\section{License and ethics}

From a license point of view, MS-COCO \cite{COCO} and ShapeNet \cite{ShapeNet} are both CC-BY 4.0 licensed while Objaverse \cite{Objaverse} is licensed ODC-BY. Then, 3D-COCO is licensed in a compatible and nonrestrictive way regarding the datasets that are used.

Regarding ethical considerations, 3D-COCO's contribution to the field extends solely to the addition of 3D CAD models and the implementation of 2D-3D alignment techniques. This augmentation of MS-COCO does not alter or affect the original dataset's adherence to privacy and ethical standards. 

\section{Conclusion}

To conclude, 3D-COCO was thought of as an extension of the original MS-COCO \cite{COCO} dataset including 27,760 3D CAD models of 80 different semantic classes collected from ShapeNet \cite{ShapeNet} and Objaverse \cite{Objaverse}. An automatic class-driven retrieval method based on IoU has been implemented to provide a 2D-3D alignment between the 860,001 training or the 36,781 validation annotations and the collected 3D models. This extension bridges the gap between MS-COCO \cite{COCO} and the 3D world: new tasks such as detection networks configurable with 3D modes, synthetic multi-view 3D reconstruction networks, or real single-view 3D reconstruction networks could be addressed thanks to 3D-COCO.

The philosophy of 3D-COCO lies in its transparency, open access, and the possibility for users to iterate over the originally proposed dataset through code sharing.

Nevertheless, for future iteration of the dataset, 3D-COCO could be improved with a better 2D-3D alignment method for articulated 3D models such as humans or animals. Exploring other retrieval methods based on neural network feature extraction or integrating new 3D models to have a more balanced number of CAD models for each class are also relevant perspectives.

\bibliographystyle{IEEEbib}
\bibliography{strings,refs}

\end{document}